\documentclass[letterpaper, 10 pt, conference]{ieeeconf}  
\IEEEoverridecommandlockouts                  
\usepackage{graphics} 
\usepackage{mathptmx} 
\usepackage{amsmath} 
\usepackage{amssymb}  
\usepackage[export]{adjustbox} 
\usepackage{bm}

\usepackage{subcaption} 
\usepackage[font=small,labelfont=bf,tableposition=top,hypcap=false]{caption} 
\usepackage{tikz}
\usepackage[compact]{titlesec}
\usepackage{xltabular}

\newcommand{\etal}{\textit{et al.~}}
  
\def\eg{\textit{e.g.},~}

\makeatletter
 \let\NAT@parse\undefined
 \makeatother
\usepackage[pdftex,
        colorlinks=true,
        bookmarks=true,
        citecolor=red,
        linkcolor=blue,
        pagebackref=false,
        pdfstartview=Fit,
        breaklinks=true
            ]{hyperref}
\usepackage{cleveref}

\title{\LARGE \bf
AcousticFusion: Fusing Sound Source Localization to Visual SLAM in Dynamic Environments
}

\author{Tianwei Zhang$^{1}$, Huayan Zhang$^{1}$, Xiaofei Li$^{*,2}$, Junfeng Chen$^{1}$, Tin Lun Lam$^{*,3,1}$, Sethu Vijayakumar$^{4,5}$

\thanks{$^{1}$Shenzhen Institute of Artificial Intelligence and Robotics for Society, the Chinese University of Hong Kong, Shenzhen}%
\thanks{$^{2}$Westlake University \& Westlake Institute for Advanced Study, Hangzhou, China.}%
\thanks{$^{3}$School of Science and Engineering, the Chinese University of Hong Kong, Shenzhen.}
\thanks{$^{4}$School of Informatics, The University of Edinburgh and The Alan Turing Institute, United Kingdom.}%

\thanks{$^{5}$The author is a visiting researcher with the Shenzhen Institute of Artificial Intelligence and Robotics for Society (AIRS).}%
\thanks{$^*$Corresponding authors: \tt\small lixiaofei@westlake.edu.cn;  \tt\small tllam@cuhk.edu.cn}}%

\begin{document}
\maketitle


\begin{abstract}
Dynamic objects in the environment, such as people and other agents, lead to challenges for existing simultaneous localization and mapping (SLAM) approaches.
To deal with dynamic environments, computer vision researchers usually apply some learning-based object detectors to remove these dynamic objects. 
However, these object detectors are computationally too expensive for mobile robot on-board processing.
In practical applications, these objects output noisy sounds that can be effectively detected by on-board sound source localization.
The directional information of the sound source object can be efficiently obtained by direction of sound arrival (DoA) estimation, but the depth estimation is difficult.
Therefore, in this paper, we propose a novel audio-visual fusion approach that fuses sound source direction into the RGB-D image and thus removes the effect of dynamic obstacles on the multi-robot SLAM system.
Experimental results of multi-robot SLAM in different dynamic environments show that the proposed method uses very small computational resources to obtain very stable self-localization results.  

\end{abstract}


\section{\textsc{Introduction}}
Visual-SLAM is a core technique for a robot to understand the external environment and perform self-orientation.
However, in real workspaces, there are many dynamic objects such as moving human talkers and other robots (in multi-robot systems). These dynamic objects disrupt most existing vision-SLAM systems: 
In the case of localization, visual odometry fails because the moving camera cannot acquire enough static visual features from the background that is obscured by dynamic objects. For environmental mapping, these moving obstacles with distorted shapes are not supposed to appear in the final map.
To deal with this dynamic environment problem, an intuitive idea is to introduce a detector for moving objects, find and remove these objects by pre-processing them in the SLAM front end, and then enable the static SLAM algorithm.
Many dynamic SLAM methods \cite{DynaSLAM,DSSLAM,zhang2018posefusion}, on the one hand, introduce learning-based object detectors to segment bounding boxes or templates of specific movable objects, such as people and vehicles.
Other methods, which fall under the motion segmentation category \cite{sf, ff, sw},  decouple dynamic pixels from static background pixels by comparing camera motion consistency --- clustering dynamic pixel points and removing them.
Whether the approach is for motion segmentation or object recognition, both frameworks expend huge computational resources for discovering and removing the dynamic components of the visual perceptual input. 
These approaches typically rely on a dedicated GPU to cope with dynamic obstacles for real-time performance, which limits their application in online, mobile robotic systems with limited computational resources.
\begin{figure}[tb]
    \centering
    \includegraphics[width=\columnwidth]{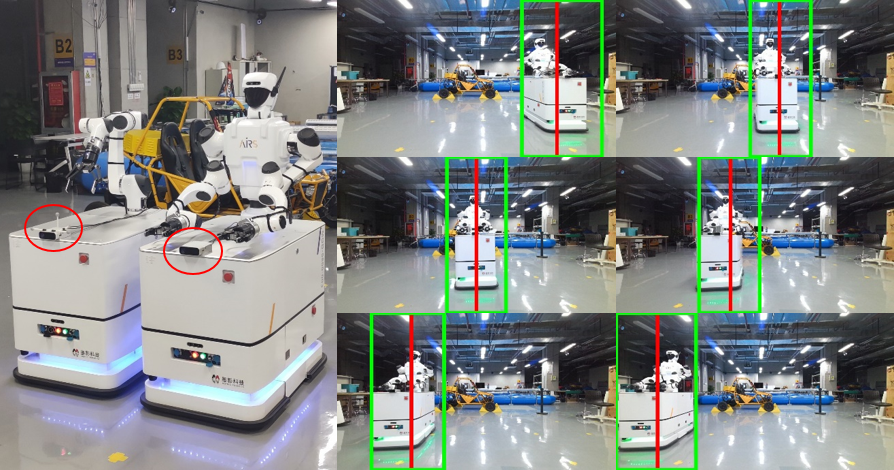}
    \caption{AcousticFusion SSL testing on the AIRS Mobile Manipulation system robots. An Azure Kinect was installed on the mobile base to keep the camera stable while the robots were moving.    
    We show six images taken by this sensor with the SSL confidence interval marked in green.}
    \label{fig:profile}
\end{figure}

Audio Human-robot interaction (HRI) requires the detection of different sound sources in the environment. This function usually requires the use of a microphone array to localize, track, and decompose the different sound sources in real-time.
In the field of SSL research, to localize and track the speakers in real-world environments, the classical methods are mostly based on estimating the time differences of arrival (TDOA) between microphones.
Knapp \etal proposed a classic TDOA estimation method in \cite{ssl1} with generalized cross-correlation. 
Chen \etal proposed a TDOA estimation framework for single-speaker localization in \cite{ssl2}. 
In the case of multiple speakers, DiBiase \etal provided a beamforming-based method named steered-response power in \cite{ssl3}, and Ishi \etal proposed a famous multiple signals classification approach in \cite{ssl4}.
Recently, In \cite{xiaofei-ssl}, Li \etal proposed a direct-path relative transfer function method combined with exponential gradient for the simultaneous localization and tracking of multiple moving speakers. This method is robust against the reverberation effect, which is especially important for indoor SLAM applications.  
Note that, limited by the compact structure of the microphone array, Sound Source Localization (SSL) mentioned here usually only estimates the direction of sound sources, and the range/depth estimation is not conducted. 

In this paper, we use the SSL method to detect sounding obstacles and mark areas of these obstacles in the image and remove them to enable visual-SLAM in the dynamic environment. This work uses the Microsoft Azure Kinect sensor. It is compact in design ($10~cm\times 12.5~cm$) and includes a microphones array and an RGB-D camera. The Azure Kinect captures asynchronous sound signals and images, and our method processes and fuses the sound signals into the images. Hence we named the proposed method AcousticFusion.
There are several advantages of fusing sound and visual signals of Azure Kinect for the mobile robot: 
1) lower on-board power and computation costs. The SSL method \eg \cite{xiaofei-ssl} performed efficient online multiple sound source azimuth detection using an on-board CPU. 
2) As shown in Fig. \ref{fig:profile}, the microphone array is small in size and low in power consumption but brings 360-degree azimuth detection and tracking capability (RGB-D cameras have a 90-degree azimuth field-of-view).
3) The Azure Kinect SDK can provide human speaker recognition and voice-to-text functions promising in HRI applications. 
In summary, this work contributes to:
\begin{enumerate}
    \item  A novel \textbf{dynamic SLAM approach} based on sound source detection and sparse feature visual odometry.   
    \item  An efficient and robust method for \textbf{fusing SSL and RGB-D image}. 
    \item  A visual odometry database with \textbf{synchronized sound and RGB-D images}.
\end{enumerate}
The database and code will be open-sourced depending on acceptance.

\section{\textsc{Related Works}}
\subsection{Dynamic visual SLAM}
Most of the existing dynamic SLAM solutions try to deal with the dynamic environment problem by finding and removing dynamic objects. Based on their object recognition approaches, 
we divide the current state-of-the-art into motion segmentation-based and object detection-based methods. 

\textbf{Object detection-based dynamic SLAM methods}
usually utilize advanced deep learning-based object detectors to remove dynamic objects and then enable the classical static SLAM frameworks in the dynamic environments.
Bescos \etal \cite{DynaSLAM} proposed DynaSLAM which applied Mask-RCNN \cite{mask-rcnn} to detect human objects in RGB images and adopted ORB-SLAM2 (ORB2) \cite{orb2} for camera tracking.
DynaSLAM performed accurate human silhouette segmentation, but it spent abound $300~ms$ per frame.  

\textbf{Motion segmentation based approaches} attempted to find dynamic pixels or point clouds rather than recognizing moving objects. 
Scona \etal proposed StaticFusion (SF) \cite{sf} that combined scene flow computation with Visual Odometry to achieve real-time static background reconstruction in a small-sized room.
Zhang \etal proposed FlowFusion \cite{ff} that utilized optical flow residuals for dynamic object segmentation and remove the dynamic point clouds for dense background reconstruction.
Judd \etal provided a multi-object motion segmentation method in \cite{mvo}, which applied sparse feature points alignment to separate and track multiple rigid objects. 
Dai \etal proposed to distinguish dynamic or static feature points using motion consistence in \cite{rgb-dySLAM}. 
All of the above dynamic SLAM solutions can not work on real-time mobile robot platforms without GPUs.

\subsection{Audio-Visual Fusion Methods}
Hospedales \etal  \cite{AVFusion2008} proposed a Bayesian model-based audio-visual fusion framework to segment, associate, and track multiple objects in audiovisual sequences.
Li \etal presented an SSL-based HRI system in \cite{xiaofei-nao}. They calibrated the sound sources' corresponding pixel coordinates. Hence an NAO robot head with four microphones performed robust azimuth localization under difficult acoustic conditions.
Ban \etal proposed an audio-visual fusion method for multi-speaker tracking in \cite{AV1}, which fused direct-path related transfer function features into the Bayesian face observation model. Then, they updated this multi-speaker tracking module in \cite{AV2}. It can track multiple speakers and locate the sound source into the bounding box of the speaker's head. In that work, CNN-based offline person detection is required, which is two frames per second (\textit{fps}) on a GTX 1070 GPU.   

The audio-visual fusion works \cite{xiaofei-nao, AV1, AV2}, all use static robots to track the observer.
For the moving robot, in \cite{acousticSLAM}, Evers \etal proposed an acoustic SLAM framework that is different from the general concept of SLAM. Acoustic SLAM applied the SSL technique passively localize a moving observer and simultaneously mapped the positions of surrounding sound sources. It does not work on robot self-localization and environment mapping. 
A recent work \cite{audioSLAM2020} proposed to use two moving microphone arrays to do SSL separately and to estimate the sound source location using the intersection of sound source direction extension lines.

\begin{figure*}[tb]
    \centering
    \includegraphics[width=\linewidth]{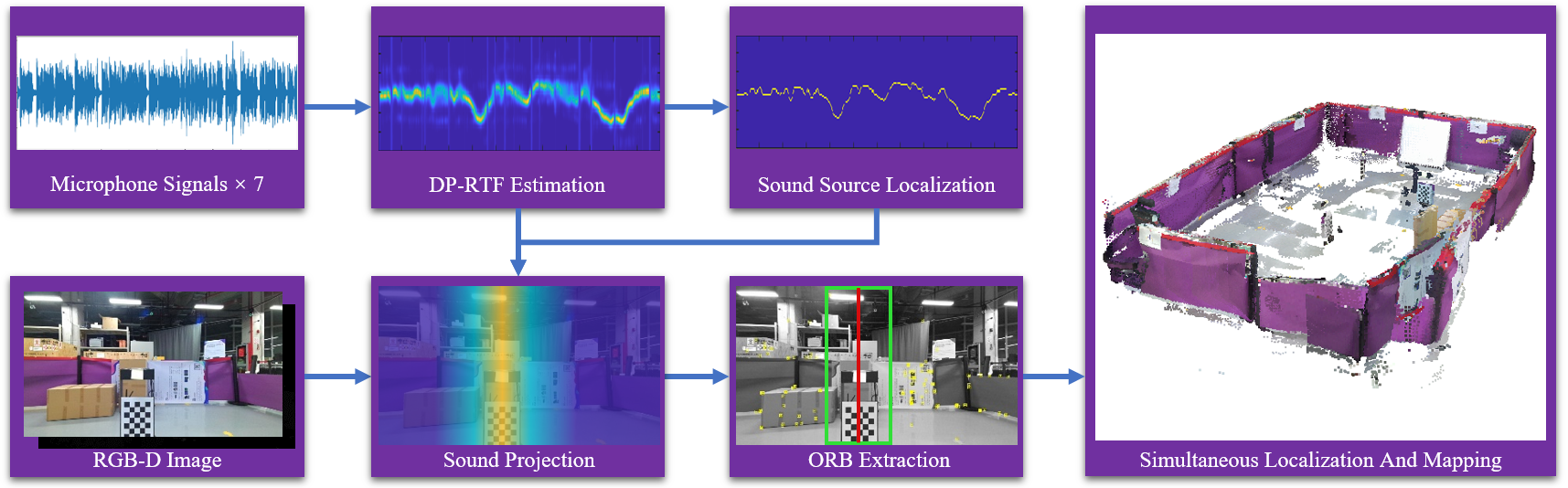}
    \caption{AcousticFusion Flowchart. Azure Kinect captures sound signals and RGB-D images. These sound signals from seven microphones are processed by feature extraction and clustering to output time-varying azimuths of the sound sources. Then, these sound source azimuth angles are fused into the image space. For the RGB-D images, we first invalidate the depth values within the sound source localization region and then extract the visual features (the yellow dots) for camera motion estimation and mapping.}
    \label{fig:flowchart}
    \vspace{-0.6cm}
\end{figure*}

\section{APPROACH}
\subsection{Overview}
Our proposed AcousticFusion framework combines SSL technology into a mobile robot vision-SLAM system. The flowchart is shown in Fig.~\ref{fig:flowchart}:
The SSL module takes the sound signals collected by seven microphones as input, and after feature extraction and clustering, outputs time-varying sound source azimuth angles, which are then fused into the image space.
For the RGB-D images acquired by the same device, we first invalidate the depth values within the SSL region and then extract the visual features for the following ego-motion estimation and mapping.
\label{sec:overview}

\subsection{Online Multiple Sound Source Localization}\label{sec:dprtf}
 \subsubsection{Extraction of Localization Feature}
In the time domain, we represent the microphone signal as: $y^m(t)= h^m(t) * x(t)$, 
where $m=1,\dots,M$ and $t$ denote the microphone and time indices, respectively. The $m$-th microphone signal $y^m(t)$ is the convolution (denoted as $*$) of the source signal $x(t)$ and room impulse response (RIR)  $h^m(t)$. In the short-time Fourier transform (STFT) domain, this convolution is represented with convolutive transfer function (CTF): $y^m_{p,k}=h^m_{p,k} * x_{p,k}$, where  $p=1,\dots,P$ and $k=0,\dots,K-1$ denote the time-frame and frequency indices, respectively. Analogous to the time-domain representation, the STFT coefficients of microphone signal $y^m_{p,k}$ is the convolution (along the time-frame axis) of the STFT coefficients of source signal $x_{p,k}$ and CTF $h^m_{p,k}$ (the STFT representation of RIR). The CTF coefficients  encode the RIR taps for one subband, and preserve the reverberation structure of RIR. The direct-path propagation presents at the first RIR taps, and thence at the first CTF coefficient.
Localization of sound source relies on estimating the direct-path propagation of the source signal to multiple microphones. In the following, we first estimate the entire CTFs based on the cross-relation method \cite{xu1995} from the microphone signals, and then extract the first CTF coefficients for sound source localization. Since the CTF estimation is independently conducted for each frequency, for notational simplicity, the frequency index $k$ will be omitted until the next section. 

 For one microphone pair $(m,n)$,  we have the cross-relation $y^m_{p}*h^n_{p} = y^n_{p}*h^m_{p}$. Let $Q$ denote the number of CTF coefficients,  $\mathbf{h}^m=[h^m_{0},\dots,h^m_{Q-1}]^T$ and $\mathbf{y}^m_{p}=[y^m_{p},\dots,y^m_{t-Q+1}]^T$ denote the vector form of CTF and microphone signal, where $^T$ denotes matrix/vector transpose. The cross-relation can be written in vector form as   
\begin{align}\label{eq:cc}
\mathbf{y}_{p}^{m \ T}\mathbf{h}^n=\mathbf{y}_{p}^{n \ T}\mathbf{h}^m.
\end{align}

The CTF vector to be estimated of all channels are concatenated as $\mathbf{h}=[\mathbf{h}^{1 \ T},\dots,\mathbf{h}^{M \ T}]^T$. To represent the pair-wise cross-relation with respect to $\mathbf{h}$, the microphone signal vectors are concatenated as: 
 \begin{align}\label{eq:xij}
\mathbf{y}^{mn}_{p} = [{0},\dots,{0},  \mathbf{y}_{p}^{n \ T}, {0},\dots,{0}, -\mathbf{y}_{p}^{i \ T}, {0},\dots,{0}]^T,
 \end{align}  
where the zero-elements are set to respond to the CTF vectors other than the $m$-th and $n$-th microphones, so that the cross-relation (\ref{eq:cc}) can be written as 
$\mathbf{y}_{p}^{mn \ T} \mathbf{h} = 0.$
There exist one trivial solution for this equation, namely $\mathbf{h}$ equals 0. To avoid this solution, the first CTF coefficient of the reference channel, say $m=r$, is constrained to be equal to $1$, namely 
 \begin{align}\label{eq:rcc}
\mathbf{y}_{p}^{mn \ T} \mathbf{h} = 0, \quad s.t. \quad h^r_{0} = 1 
 \end{align}
This can be realized by dividing $\mathbf{h}$ by $h^r_{0}$, which yields a new equation $\mathbf{y}_{p}^{mn \ T}\mathbf{h}/h^r_{0} = 0$. Moving the constant term from the left side to the right side, we have  
\begin{align}\label{eq:nor-rcc}
 \tilde{\mathbf{y}}_{p}^{mn \ T} \tilde{\mathbf{h}} = z^{mn}_{p},
 \end{align}
where $\tilde{\mathbf{y}}_{p}^{mn}$ is $\mathbf{y}_{p}^{mn}$ with the entry corresponding to $h^r_{0}$ removed, and $-z^{mn}_{p}$ is such entry. The new variable $\tilde{\mathbf{h}}$ is $\mathbf{h}$ with $h^r_{0}$ removed, and then divided by $h^r_{0}$. In $\tilde{\mathbf{h}}$, the elements corresponding to the first CTF coefficient of multiple microphones (other than the $r$-th microphone), i.e. $h^m_{0}/h^r_{0}, m\neq r$, represent the ratio between the direct-path transfer function of two microphones, and are referred to as direct-path relative transfer functions (DP-RTFs). DP-RTFs encode the localization cues, namely the inter-channel phase/magnitude difference of the direct-path signal propagation.
 
Sound source localization amounts to estimate the DP-RTFs by solving the linear problem Eq. (\ref{eq:nor-rcc}). We note that Eq. (\ref{eq:nor-rcc}) is defined for one microphone pair at one time frame. For online processing, we receive the microphone signals $\tilde{\mathbf{y}}_{p}^{mn}$ and $z^{mn}_{p}$ frame by frame, and accordingly the estimate of $\tilde{\mathbf{h}}$ will also be updated frame by frame. For one frame, all the $I=M(M-1)/2$ distinct microphone pairs are utilized. For notational convenience, we use $i=1,\dots,I$ denote the index of microphone pair to replace ${mn}$. Define the fitting error of (\ref{eq:nor-rcc}) as $e^{i}_{p} = \tilde{\mathbf{y}}_{p}^{i \ T} \tilde{\mathbf{h}}-z^{i}_{p}$. At one current frame $p$,  exploiting the microphone pairs up to $i$, online processing aims to minimize 
\begin{align}\label{eq:cost}
 J^i_{p} = \sum_{p'=1}^{p-1}\lambda^{p-p'}\sum_{i'=1}^I |e^{i'}_{p'}|^2+\sum_{i'=1}^i |e^{i'}_{p}|^2,
 \end{align}   
which sums up the fitting error of all the currently available frames and microphone pairs. 
Along with the increase of $p$ or $i$, this error is recursively updated with one new error term, i.e. $|e^{i}_{p}|^2$, for which $\tilde{\mathbf{h}}$ can be efficiently estimated with the recursive least squares algorithm (please find more details from \cite{xiaofei-ssl}). At each frame $p$, one estimate of $\tilde{\mathbf{h}}$ is obtained, denoted as $\hat{\mathbf{h}}_{p}$. For the dynamic case (either speaker or microphone array is moving), $\tilde{\mathbf{h}}$ is time-varying, and the estimate $\hat{\mathbf{h}}_{p}$ reflects the current value of $\tilde{\mathbf{h}}$. To catch up the variation of $\tilde{\mathbf{h}}$, the older frames are exponentially forgotten by the the forgetting factor $\lambda \in (0,1]$. This factor can be set to 1 for the static case in which $\tilde{\mathbf{h}}$ is constant. 

Up to now, we consider the noise-free single-speaker case. To suppress noise, the inter-frame spectral subtraction algorithm proposed in \cite{xiaofei-2015icassp} can be easily integrated into the current framework. As for the multiple-speaker case, the W-disjoint orthogonality assumption \cite{yilmaz2004} is  used, which assumes that the speech signal is dominated by only one speaker in each small region of the STFT domain, because of the natural sparsity of speech signals in this domain. Based on this assumption, the CTF estimates (with frequency index $k$ added), i.e. $\hat{\mathbf{h}}_{p,k}$, belongs to only one of the multiple speakers. Finally, at frame $p$, from $\hat{\mathbf{h}}_{p,k}$, we extract the DP-RTFs as localization features, denoted as ${a}^m_{p,k}, \ m\in[1,M], m\neq r;k\in[0,K-1]$, and each feature is associated with a single speaker. Note that those time-frequency bins dominated by noise or multi-speaker are not used for the following localization step.

\subsubsection{Feature Clustering for Localization} \label{sec:eg}

The complex Gaussian mixture model is used to cluster the features to each active speaker. 
Each component of the mixture model is set to represent one candidate source location. Let $d=1,\dots,D$ and $w^d$ ($w^d \geq 0$ and $\sum_{d=1}^Dw^d=1$) denote the $d$-th candidate location and the prior probability of the $d$-th mixture component, respectively. The probability, that one feature ${a}^m_{p,k}$ is emitted
by candidate locations, is the mixture of complex Gaussian probabilities: 
\begin{equation}\label{eq:CGMM}
P({a}^m_{p,k}) = \sum_{d=1}^Dw^d\mathcal{N}_c({a}^m_{p,k};\bar{a}_k^{m,d},\sigma^2),
\end{equation}
where the mean $\bar{a}_k^{m,d}$ is the constant theoretical DP-RTF, which can be precomputed using the theoretical model of the signal's direct-path propagation. The variance $\sigma^2$ is empirically set as a constant value. The prior probability (weight) $w^d$ is the only free model parameter, and can be estimated by maximizing the log-likelihood of all the available features, namely
\begin{align}\label{eq:loglik}
\mathop{\textrm{max}}_{w^d,\ d=1,\dots,D} \sum_{{a}^m_{p,k}}\text{log}(P({a}^m_{p,k})).
\end{align}
This likelihood maximization problem can be easily solved by the well-known expectation-maximization algorithm. For the dynamic case, $w^d$ is time-varying (thence denoted as $w_p^d$), and can be estimated with recursive expectation-maximization algorithm. 
The optimized weight $w_p^d$ represents the probability that an active speaker is present at the $d$-th candidate location. Sound source localization can be conducted by detecting the peak of $w_p^d$ along the $d$ axis. 

In this work, we use the microphone array embedded on an Azure Kinect to conduct SSL. The topology of the microphone array is shown in \ref{fig:projection}, which is composed of seven microphones arranged in a 2D plane. The microphone array is placed to be parallel to the horizontal plane, which is thus suitable to perform horizontal (azimuth) localization. A total of $D=72$ candidate azimuth angles are set with 5 degrees gap between two adjacent angles to cover the whole 360 degrees azimuth space.


\subsection{Audio-Visual Data Association}
\begin{figure}[tb]
    \centering
    \includegraphics[width=\columnwidth]{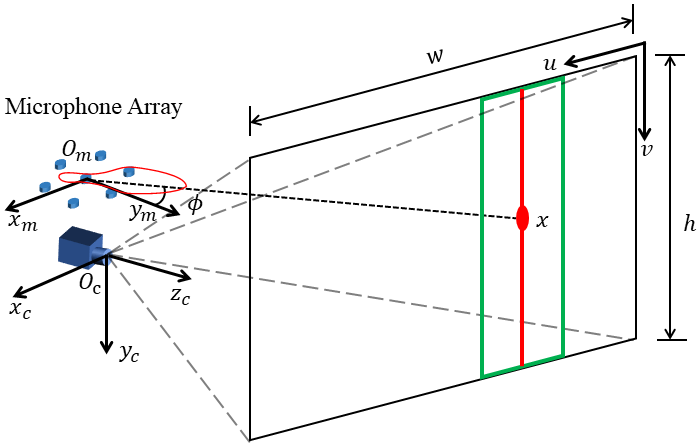}
    \caption{
    Projecting the source direction onto the image plane: We first obtain the source azimuth $\phi$ from the SSL peak, then warp the camera image to the microphone frame and extend the azimuth ray to intersect this image plane. The location of the sound source is located on the vertical line through this intersection point $x$.
    }
    \label{fig:projection}
\end{figure}

In \cite{xiaofei-nao}, Li \etal provided an audio-visual dataset that contains sound source directions that correspond to image pixels. 
They obtained these correspondences by manually labeling the loudspeaker's positions in the image. This audio-visual information association method is not robust to reverberation condition changing. The sound source to pixel correspondence changes When the robot moves.
For the mobile robot SLAM, we should update these audio-visual information correspondence time by time. 

\begin{figure*}[tb]
    \centering
    \includegraphics[width=0.98\linewidth]{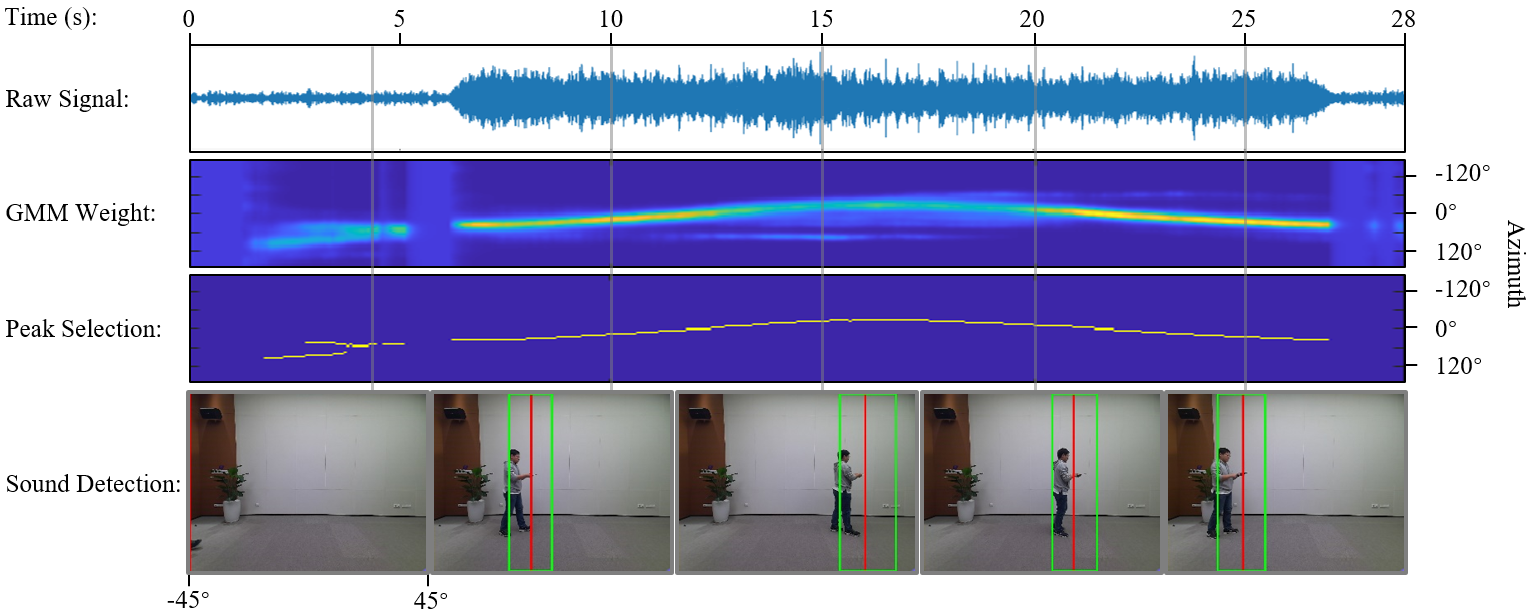}
    \caption{An example of audio-visual integration. In this scene, a pedestrian is playing a song with his cell phone. The first row is the original sound signal of a microphone channel. After sound feature extraction and clustering, the second and third rows are the results of SSL output, i.e., Gaussian probability weights and directional peaks. The song starts to play from 6.4 and ends at 26.7 seconds, and the two peak curves before 5 seconds are footsteps. The red line is the projection of the sound source peak on the image, and the width of the dynamic object bounding box in green is determined by the GMM weights.}
    \label{fig:cellphone}
    \vspace{-0.4cm}
\end{figure*}

Following the RGB-D SLAM method \cite{sf}, given two image frames: camera image frames C and microphone image frame M, at the sound sampling frame $p$ a pixel $x_C^p$ in frame C can be warped to frame M by:
\begin{align}
    \mathbf{x}_M^p = W(\mathbf{x}_C^p, T(\xi), D_C)
    \label{equ::warp}
\end{align}
where the image warping function $W$ is given by:
\begin{align}
    W(\mathbf{x}^p, T, D) = \pi(T\pi^{-1}(\mathbf{x}^p, D(\mathbf{x}^p))) \ .
    \label{equ:W}
\end{align}
$\mathbf{x}$ represents a pixel in the 2D image, $D(\mathbf{x})$ is the depth of pixel $x$. The projection function $\pi: \mathbb{R}^3 \rightarrow \mathbb{R}^2$ projects 3D points onto the image plane using the camera intrinsic matrix. The extrinsic matrix $T(\xi) \in SE(3)$ between the camera frame and microphone frame is computed using device hardware parameters. 

As the SSL module output \textit{fps} is much higher than the RGB-D camera frame rate, to update the audio-visual correspondence,
for each camera image, we warp its pixels  $x_p \in C$ to the microphone frame M using Eq. \ref{equ::warp}, label the newest estimated sound source azimuth on M, and warp the labeled pixel back to the camera frame.
For instance, in Fig.~\ref{fig:projection}, assume there is only one sound source $d_p$, fetch its sound source azimuth  $\phi\in (-180, 180]$ (the direction of the protrusion of the red circle) from $\mathcal{D}_{p}$, 
extend the azimuth ray to intersect the warped image plane and take the intersection point $x_M^p=(\mu^{'}, v^{'})$.
Then the sound source location should belong to the vertical line through $x_M^p$, so as $x_C^p$. 
However, according to the warping function Eq.~\ref{equ:W}, the unknown $D_w$ is necessary to warp $x_M^p$ back to $x_C^p$. 
In real cases, the distance of the sound source is always further than $1 m$, much larger than the distance from the center of the microphone array to the optical center of the camera ($7.4 cm$). Thus in this work, we use $D_C$ instead $D_M$.

Note that the image FOV of the camera is only 90 degrees, which is smaller than the 360 degrees of the SSL module. Therefore, when the SSL result exceeds the camera FOV, the fused sound source red line may be on the left or right border of the image, as shown in the bottom left figure of Fig.~\ref{fig:cellphone}. This red line labels the detected sound source objects, which will be removed as exceptions in the following SLAM visual feature extracting phase.  

\subsection{Visual Feature Extraction and Visual Odometry}
\begin{figure}[ht]
  \begin{minipage}[t]{0.5\linewidth}
    \centering
    \includegraphics[width=4cm]{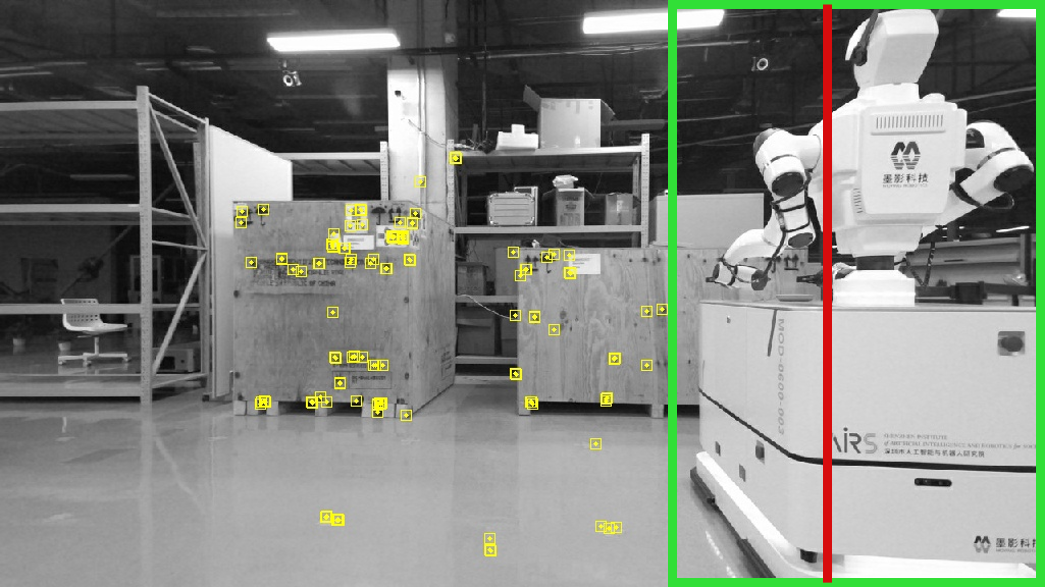} 
  \end{minipage}%
  \begin{minipage}[t]{0.5\linewidth}
    \centering 
    \includegraphics[width=4cm]{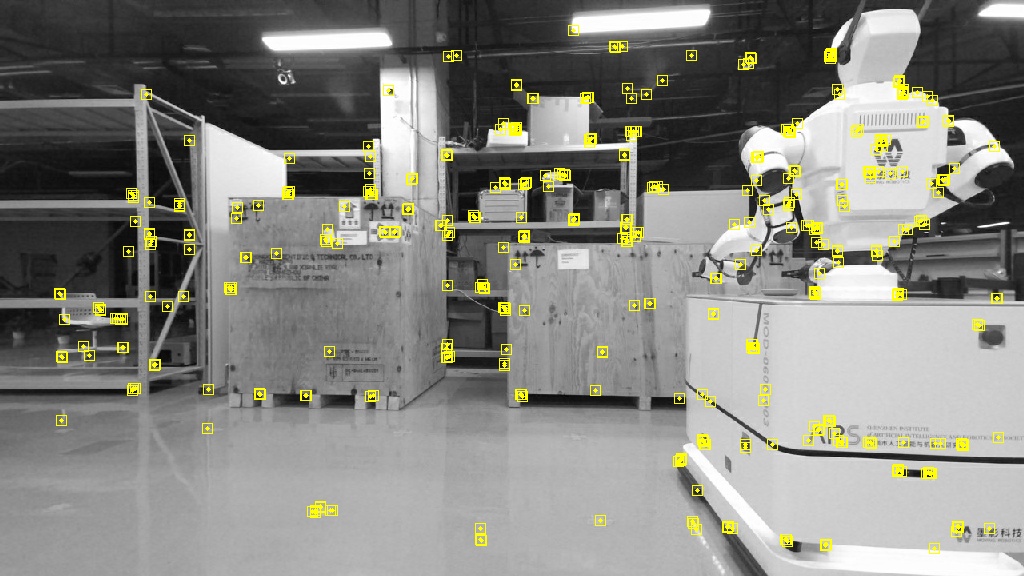} 
  \end{minipage}
  \begin{minipage}[t]{0.5\linewidth} 
    \centering
    \includegraphics[width=4cm]{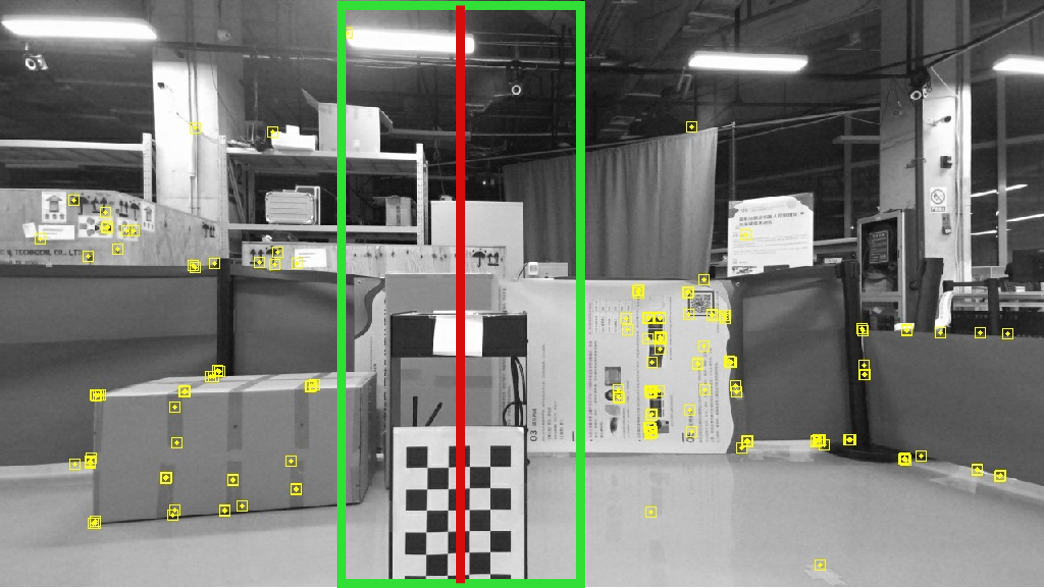} 
    \caption*{(a) Ours} 
  \end{minipage}%
  \begin{minipage}[t]{0.5\linewidth}
    \centering 
    \includegraphics[width=4cm]{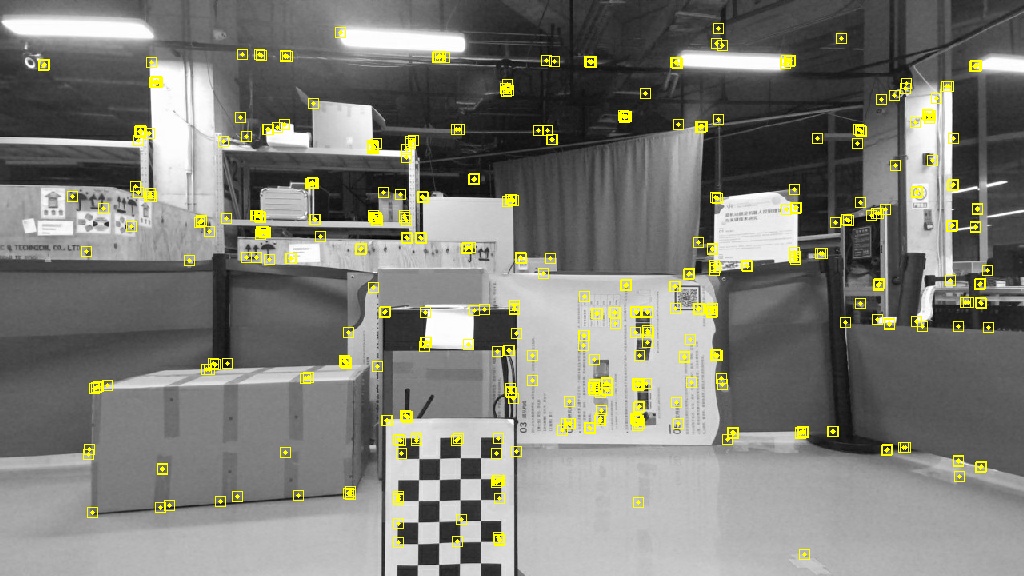}
    \caption*{(b) Openvslam} 
  \end{minipage}
\caption{Visual feature extraction in multi-robot scene. We set the width of the feature-free area according to the SSL result. In (a), our method only extracts features from the static backgrounds. In (b), Openvslam extracts wrong environmental features from the other moving robot surfaces.}
\label{fig:orb-feature}
\end{figure}

The SSL module outputs the weights (probabilities) of candidate azimuth angles associated with active speakers, and the localized sound source directions obtained by detecting the peaks of weights. 
With the sound source direction as the center, we grab an image region to cover the dynamic obstacle as much as possible according to the weights of candidate azimuth angles. 

 Sound source directions are estimated by detecting the peak of GMM weights $w_p^d$ along the $d$ axis. The candidate locations corresponding to the peaks of $w_p^d$ are denoted as $d_{p,j}\in[1,D],\ j=1,\dots,J$, where $J$ denotes the number of detected sound sources. To cover the whole visual obstacles, we need to estimate the obstacle regions in the image. The center of obstacle regions are set to be the sound source directions $d_{p,j}\in[1,D],\ j=1,\dots,J$. The region boundaries are separately determined for each visual obstacle. To determine each of the left and right region boundaries for sound direction $d_{p,j}$, for example the right boundary $b_{p,j}^{\text{right}}$, the GMM wights $w_p^d$ are checked one by one from $d_{p,j}+1$ to its right candidates until the following condition is satisfied: 
 \begin{align}
 w_p^{d+1}\ge w_p^{d} \quad \text{or} \quad w_p^{d+1}<\delta w_p^{d_{p,j}},
 \end{align}
 then $b_{p,j}^{\text{right}}$ is set to $d$. This condition means i)   the weight cannot increase, as the increasing weight indicates the emerging of a new sound source; ii) the weight should not be smaller than $\delta w_p^{d_{p,j}}$, where $0\leq\delta\leq1$ is empirically set to reflect our prior knowledge about the size of the visual obstacle.  Finally, at frame $p$ and for sound source $j$, the obstacle region is represented by the region center ${d_{p,j}}$ (red line as shown in Fig. \ref{fig:profile}, \ref{fig:flowchart}, \ref{fig:projection}, \ref{fig:cellphone} and \ref{fig:orb-feature}), and the region boundaries $\{b_{p,j}^{\text{left}},b_{p,j}^{\text{right}}\}$ (green bounding box as shown in Fig. \ref{fig:profile}, \ref{fig:flowchart}, \ref{fig:projection}, \ref{fig:cellphone} and \ref{fig:orb-feature}). These time-varying image regions track the multiple moving visual obstacles and will be adopted for the following SLAM step.  

For example, in the flowchart Fig. \ref{fig:flowchart}, after the azimuth projection of the sound source, we compute this region (green bounding box) in the RGB-D frame and invalidate the depth values in this region in the Depth image.
Then, we extract ORB visual features on this pair of RGB and Depth images so that the obtained features avoid moving object regions to ensure visual odometry robustness in these dynamic scenes. Thereafter, the loop detection and mapping tasks are done using Openvslam \cite{openvslam}.

\section{\textsc{Experiments and Evaluations}}
\label{sec:evaluation}

\begin{figure*}[ht]
\centering
\footnotesize
  \begin{tabular}{m{4cm}<{\centering}m{4cm}<{\centering}m{4cm}<{\centering}m{4cm}<{\centering}}
      Sp1 & Sp2 & Sp3 & Sp4 \\
    \includegraphics[height=3.0cm]{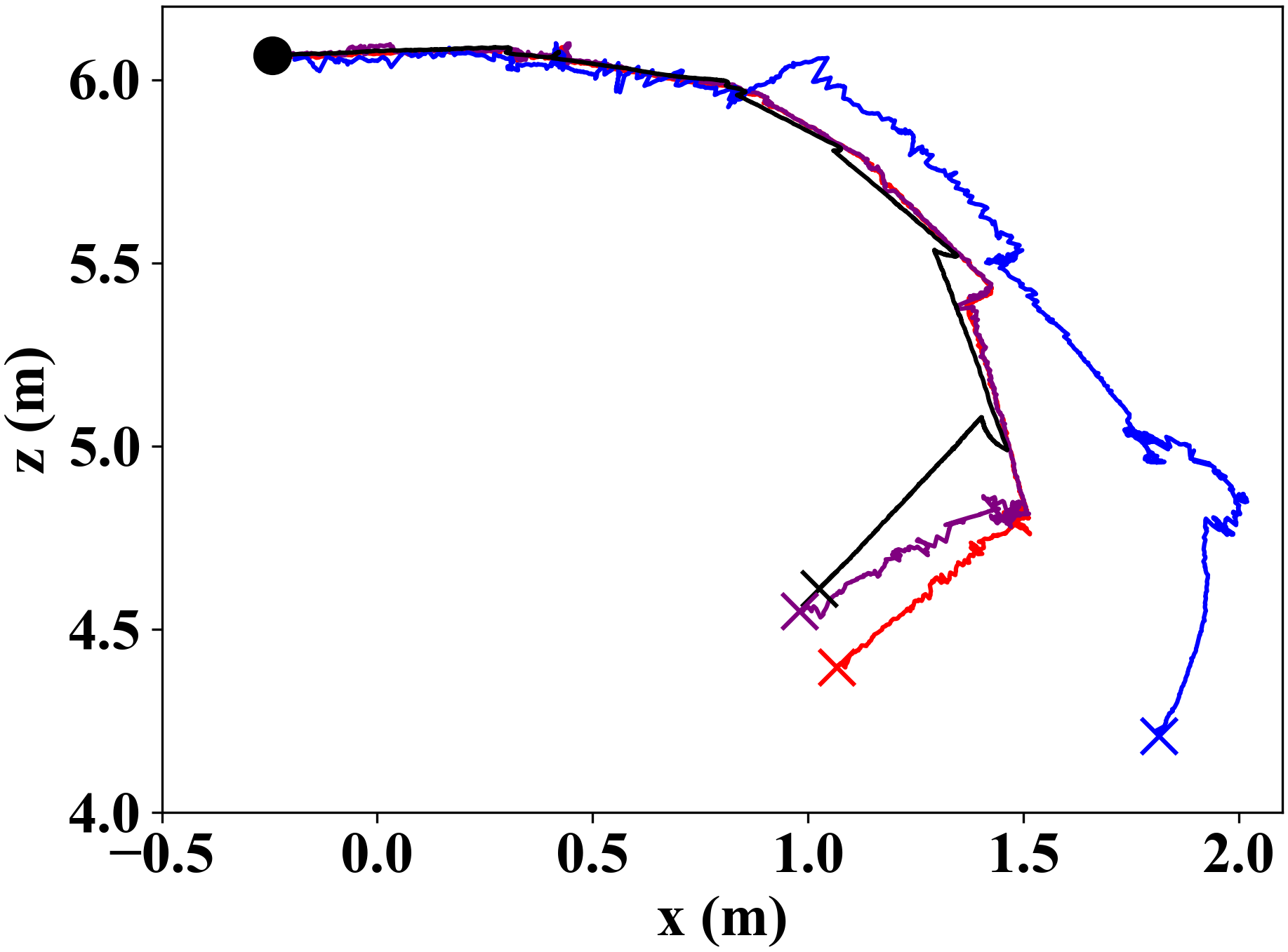} &
    \includegraphics[height=3.0cm]{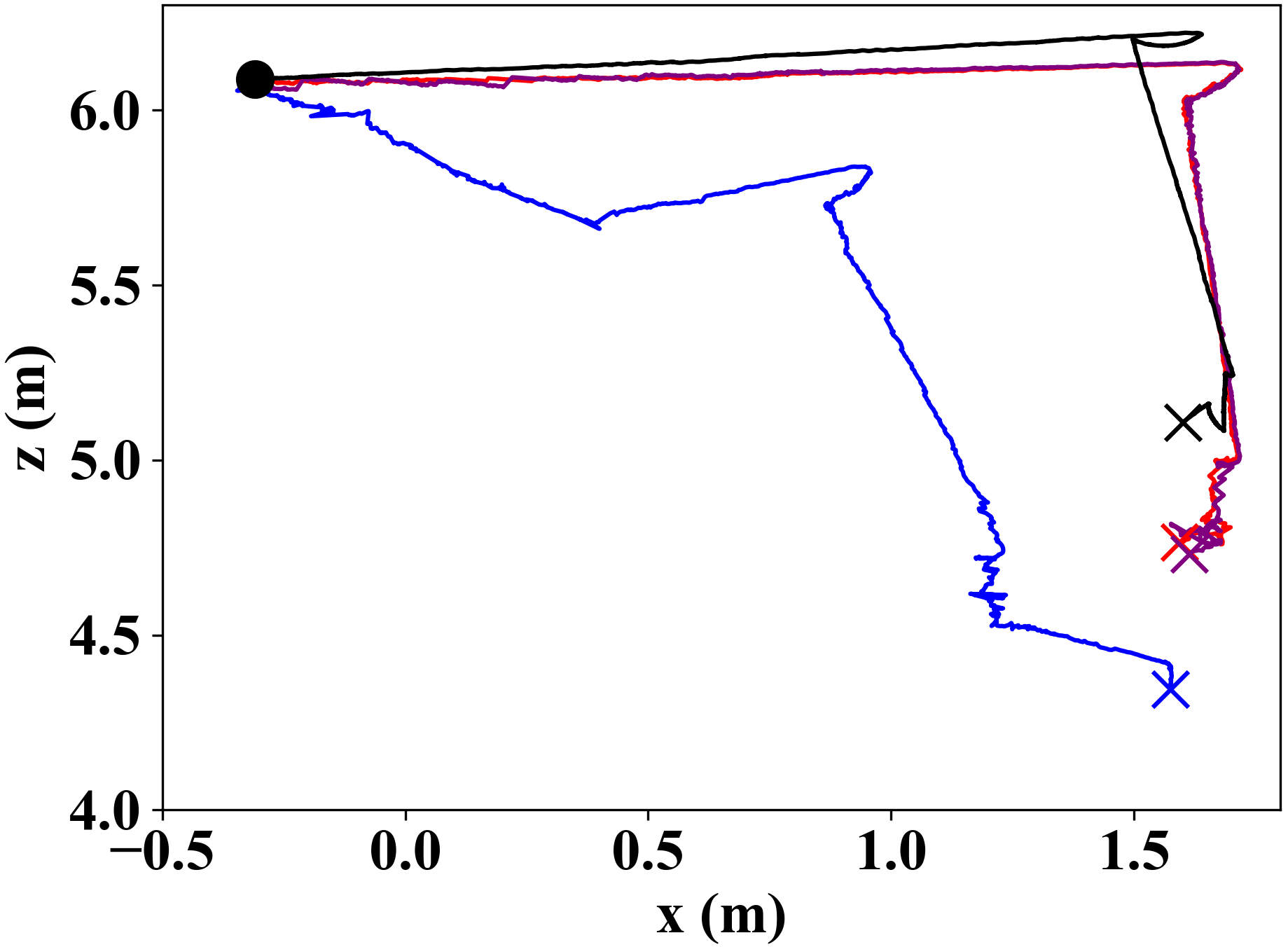} &
    \includegraphics[height=3.0cm]{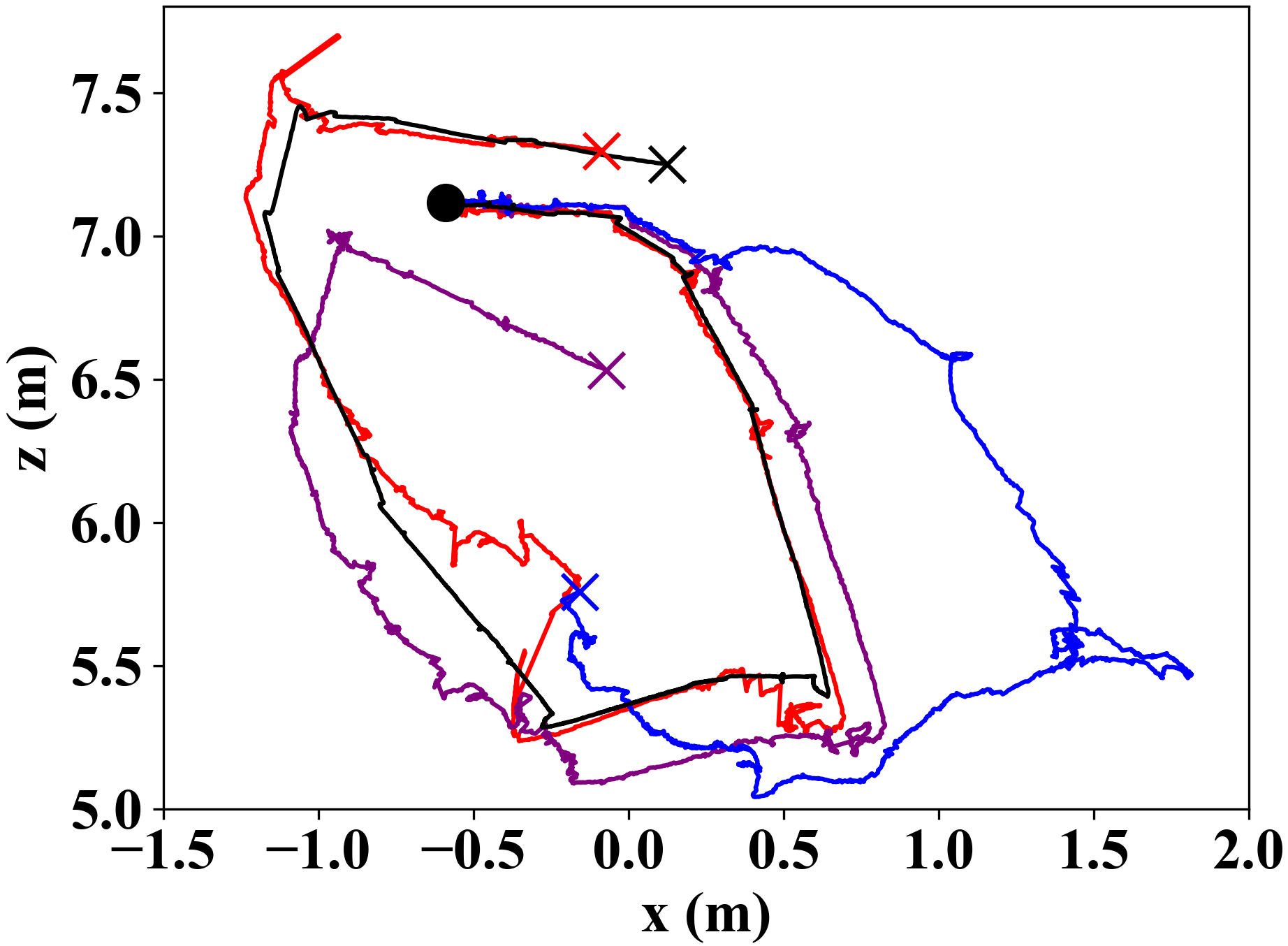} &
    \includegraphics[height=3.0cm]{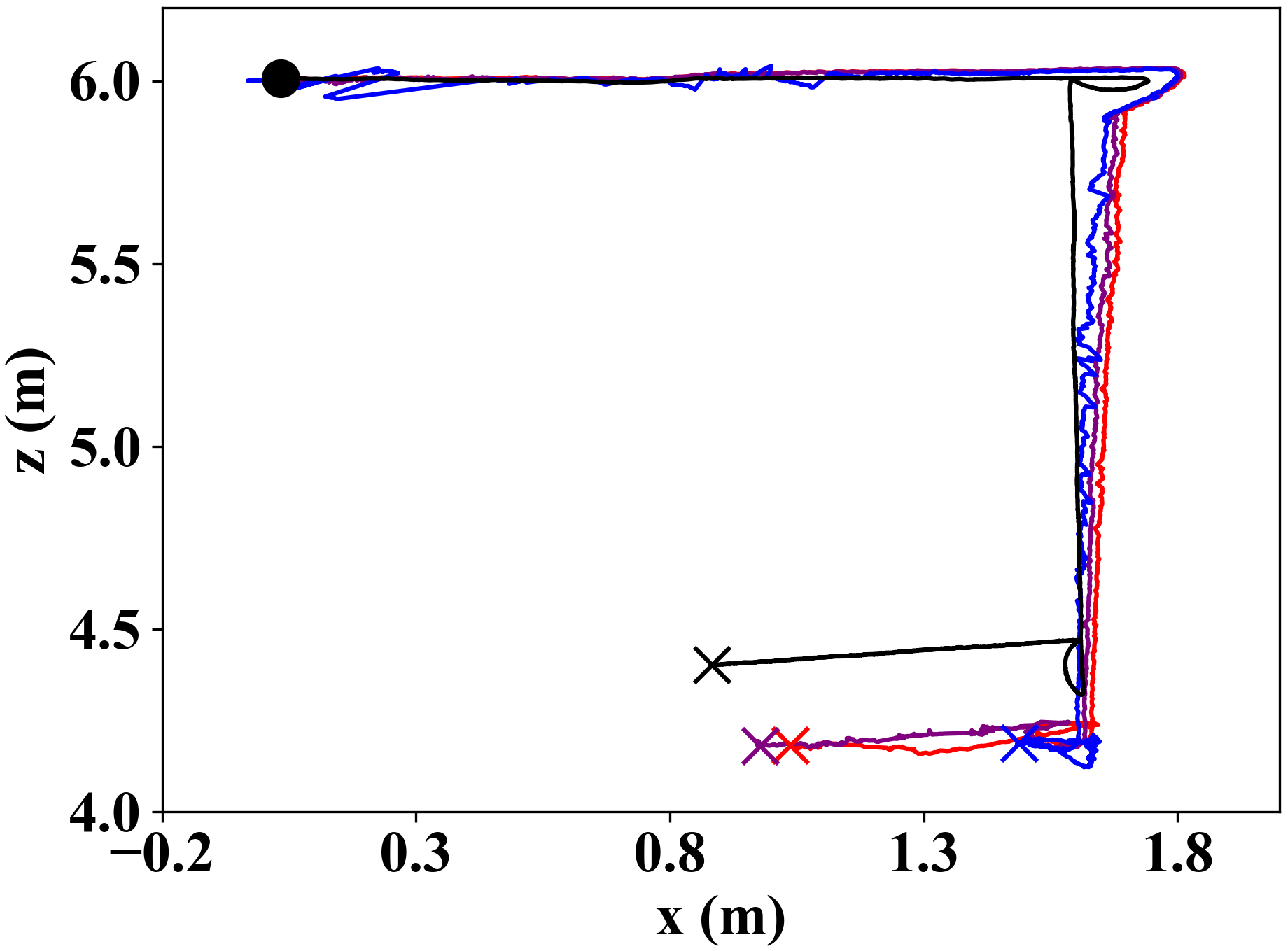} \\
    Sp5 & Mo2 & Mo3 & Mo4 \\
    \includegraphics[height=3.0cm]{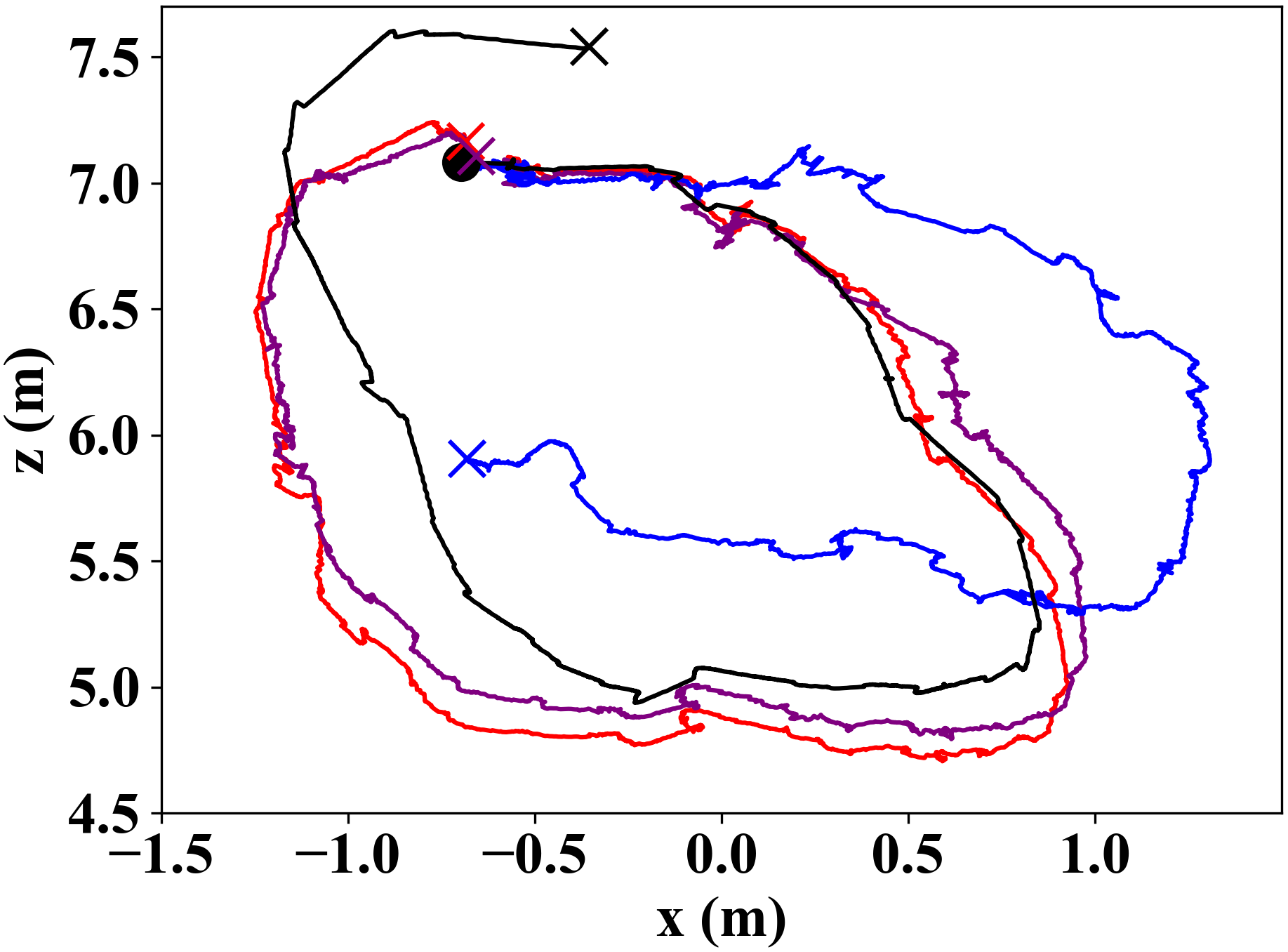} &
    \includegraphics[height=3.0cm]{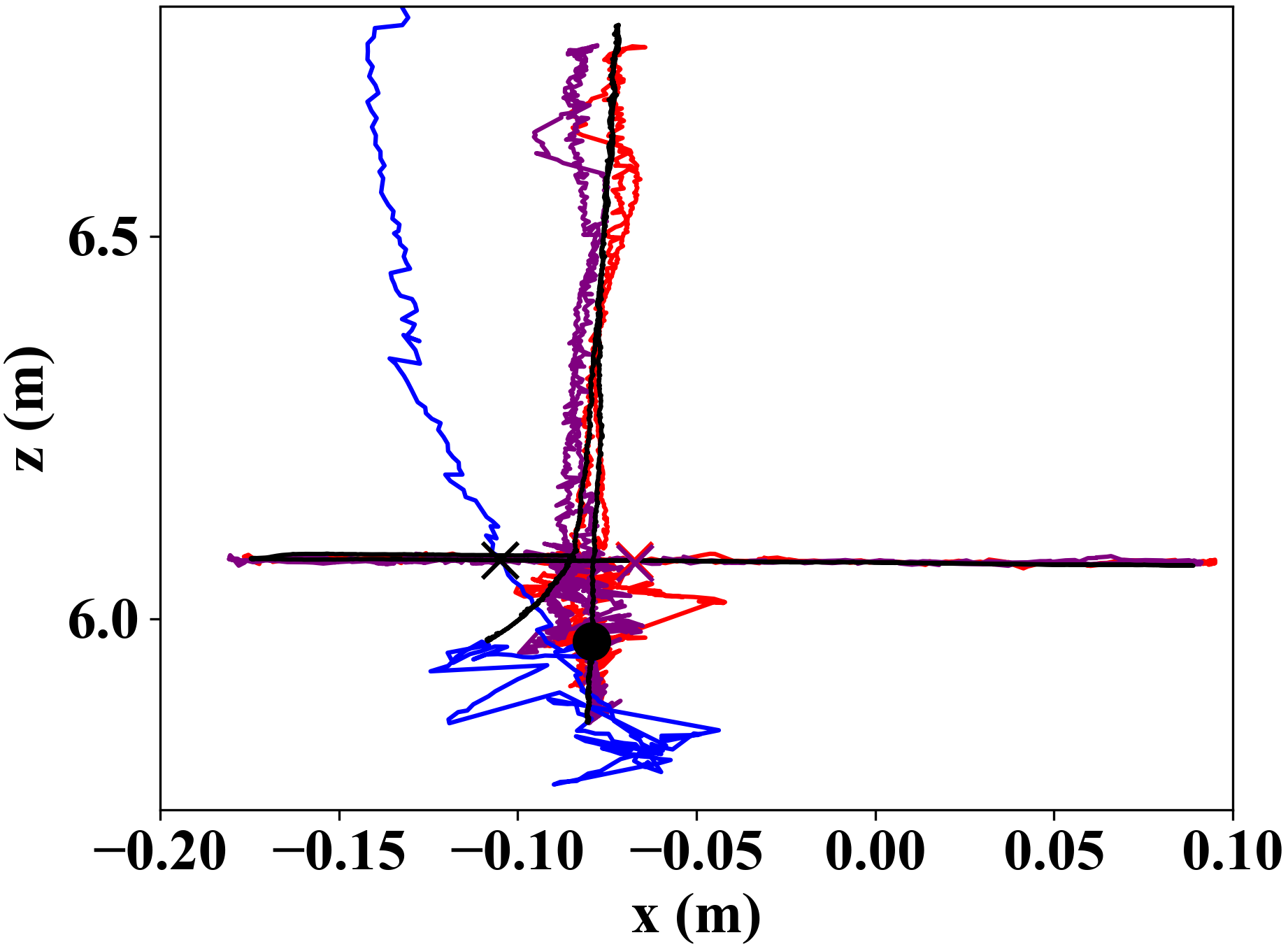} &
    \includegraphics[height=3.0cm]{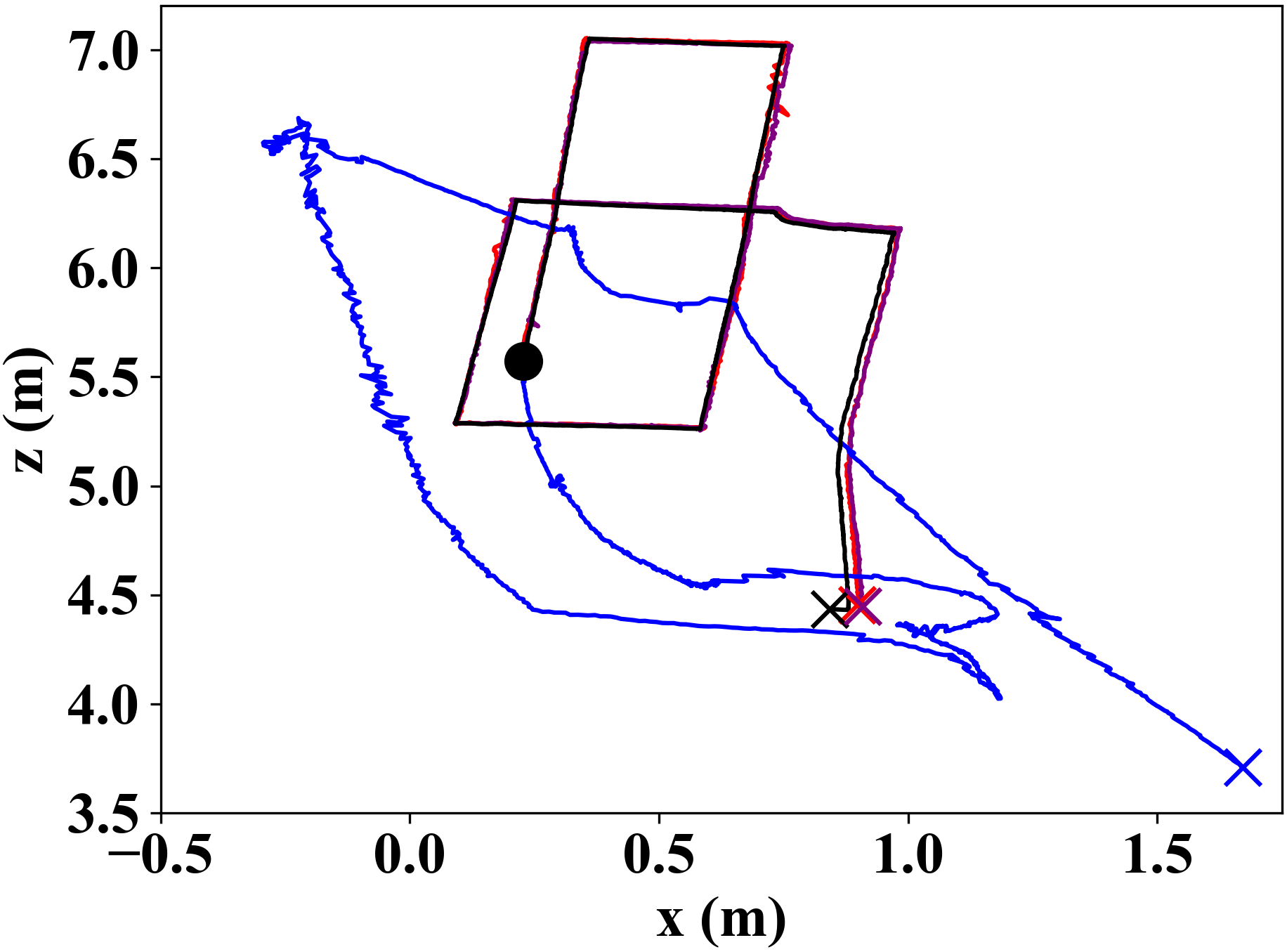} &
    \includegraphics[height=3.0cm]{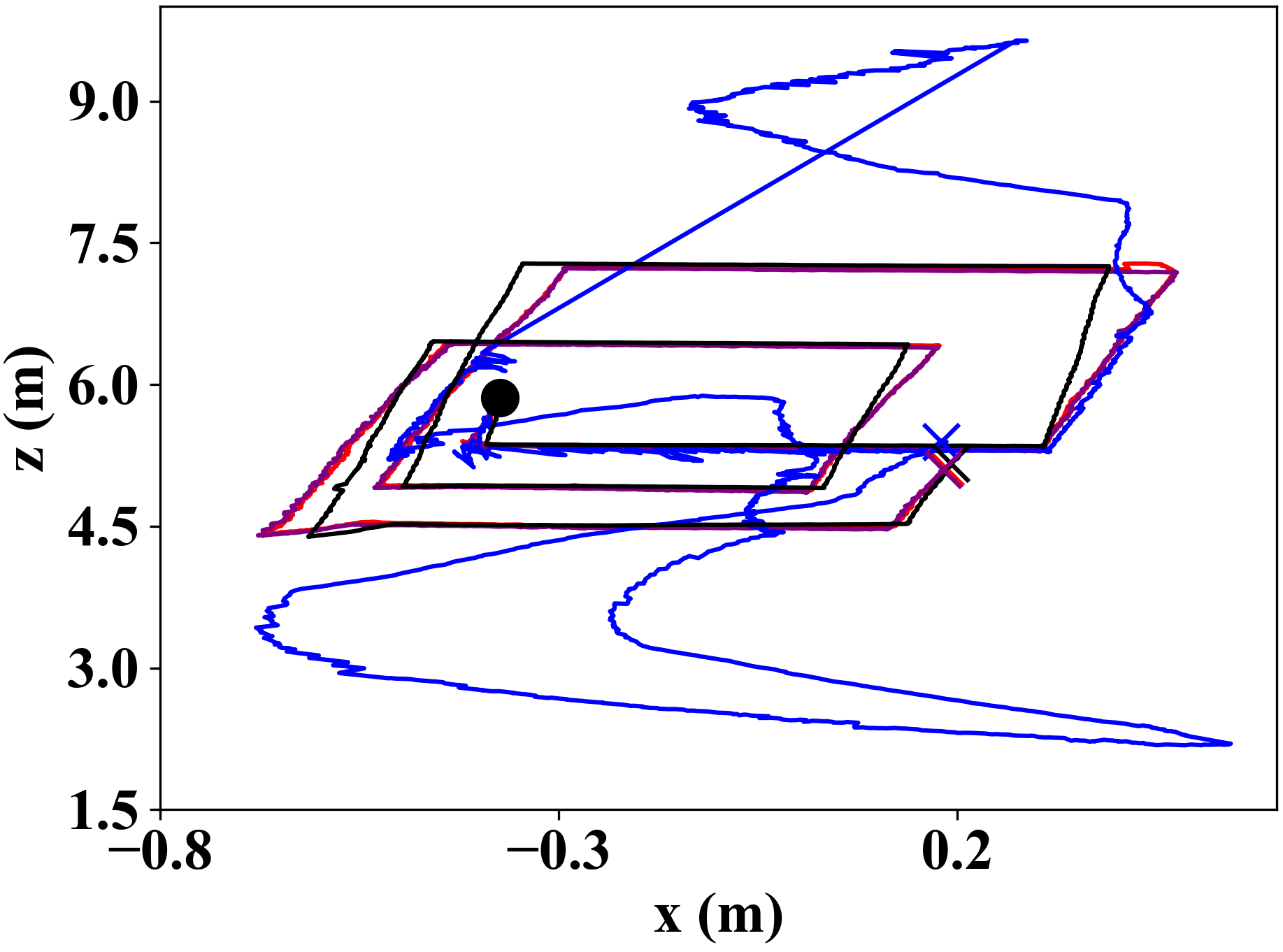} \\
    \multicolumn{4}{c}{\includegraphics[height=0.6cm]{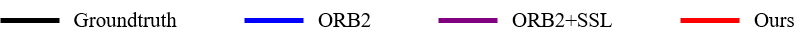}}\\
  \end{tabular}
  \caption{The plotted trajectories of Spark-T and AIRS Mobile robots sequences. 
  Our method obtains the smallest ATE, and accordingly the red curves are closest to the black ground truth curves.}
\label{fig:cam-trajectories}
\end{figure*}

In this paper, the proposed method was tested on a laptop with Intel Core$^{TM}$  i7-10875H CPU @ 2.30 GHz $\times$ 8, 64 GB System memory. In StaticFuison \cite{sf}'s comparison experiments, a GeForce RTX 2080 Ti GPU was used.
\subsection{Sound Source Localization Results}

As already partially mentioned above, $D=72$ azimuth directions at every 5 degrees in $(-180, 180]$ degrees are used as candidate directions to perform 360 degrees azimuth localization. The sampling rate of sound signals is 16,000 Hz. The STFT has a window length of $256$ samples and a hop size of $128$ samples, correspondingly SSL has an output rate of 125 Hz. The CTF length $Q$ is set to 8.   
Fig. \ref{fig:profile} and \ref{fig:cellphone} indicate the SSL module performance when the robot is static. The localization effect is stable when the target is within $3~m$. Beyond $3~m$ the localization error increases with distance. 

In addition to the sound source direction, the estimation results of the sound source area have a significant impact on the subsequent visual odometry. In multi-robot SLAM experiments, the sound source area width should be adjusted appropriately according to the size of the robot because large-sized moving targets obscure more pixels at the same distance. \eg in Fig. \ref{fig:orb-feature}, To avoid extracting undesired visual features from dynamic object surfaces, the width of the sound source area was set to 10 and 20 degrees for the Spark-T and AIRS Dual Arm Mobile Manipulation robots, respectively. 
Also, the man in Fig. \ref{fig:cellphone} appeared outside the bounding box. Because in that scene, the sound source was the phone that was farther from the center of his body.

\begin{table}[t]
\centering  
\caption{Dynamic Environment SLAM ATE RMSE (m)}
\label{T:ate}  
\begin{tabular}{p{1.2cm}m{1.2cm}<{\centering}m{0.7cm}<{\centering}m{1.8cm}<{\centering}m{0.7cm}<{\centering}}
\hline\noalign{\smallskip}
\textbf{Sequence} & ORB2  & SF & ORB2+SSL & Ours\\
\noalign{\smallskip}\hline\noalign{\smallskip}
\multicolumn{5}{c}{Spark-T Robot}\\
\noalign{\smallskip}\hline\noalign{\smallskip}
      Sp1 & 0.24  & 3.4 & 0.079 & 0.1\\
      Sp2 & 0.23  & 12.8 & 0.088 & 0.078\\
      Sp3 & 0.35  & 26.47 & 0.22 & 0.14\\
      Sp4 & 0.2  & 3.69 & 0.12 & 0.13\\
      Sp5 & 0.25  & 2.35 & 0.2 & 0.19\\
\noalign{\smallskip}\hline\noalign{\smallskip}
\multicolumn{5}{c}{AIRS Mobile Manipulation Robot}\\
\noalign{\smallskip}\hline\noalign{\smallskip}
      Mo1 & 1.32 & 14.2 & 0.13 & 0.12\\
      Mo2 & 1.59 & 18.28 & 0.089 & 0.088\\
      Mo3 & 0.94 & 0.71 & 0.038 & 0.04\\
      Mo4 & 1.45 & 1.53 & 0.18 & 0.18\\
\noalign{\smallskip}\hline\noalign{\smallskip}
\end{tabular}
\end{table}

\begin{table}[t]
\centering  
\caption{Time Cost Evaluation}
\label{T:time}  
\begin{tabular}{p{3cm}m{1.4cm}<{\centering}m{1.4cm}<{\centering}}
\hline\noalign{\smallskip}
\textbf{Method} & \textit{fps}  & GPU \\
\noalign{\smallskip}\hline\noalign{\smallskip}
      ORB2 \cite{orb2} & 26  & $\times$  \\
      DynaSLAM \cite{DynaSLAM} & 0.3 &\checkmark  \\
      SF \cite{sf} & 17 & \checkmark \\
      Ours & 14  &$\times $ \\
\noalign{\smallskip}\hline\noalign{\smallskip}
\end{tabular}
\end{table}

\begin{figure*}[ht]
\centering
\footnotesize
\begin{tabular}{m{4cm}<{\centering}m{4cm}<{\centering}m{4cm}<{\centering}m{4cm}<{\centering}}
    \includegraphics[height=2.4cm]{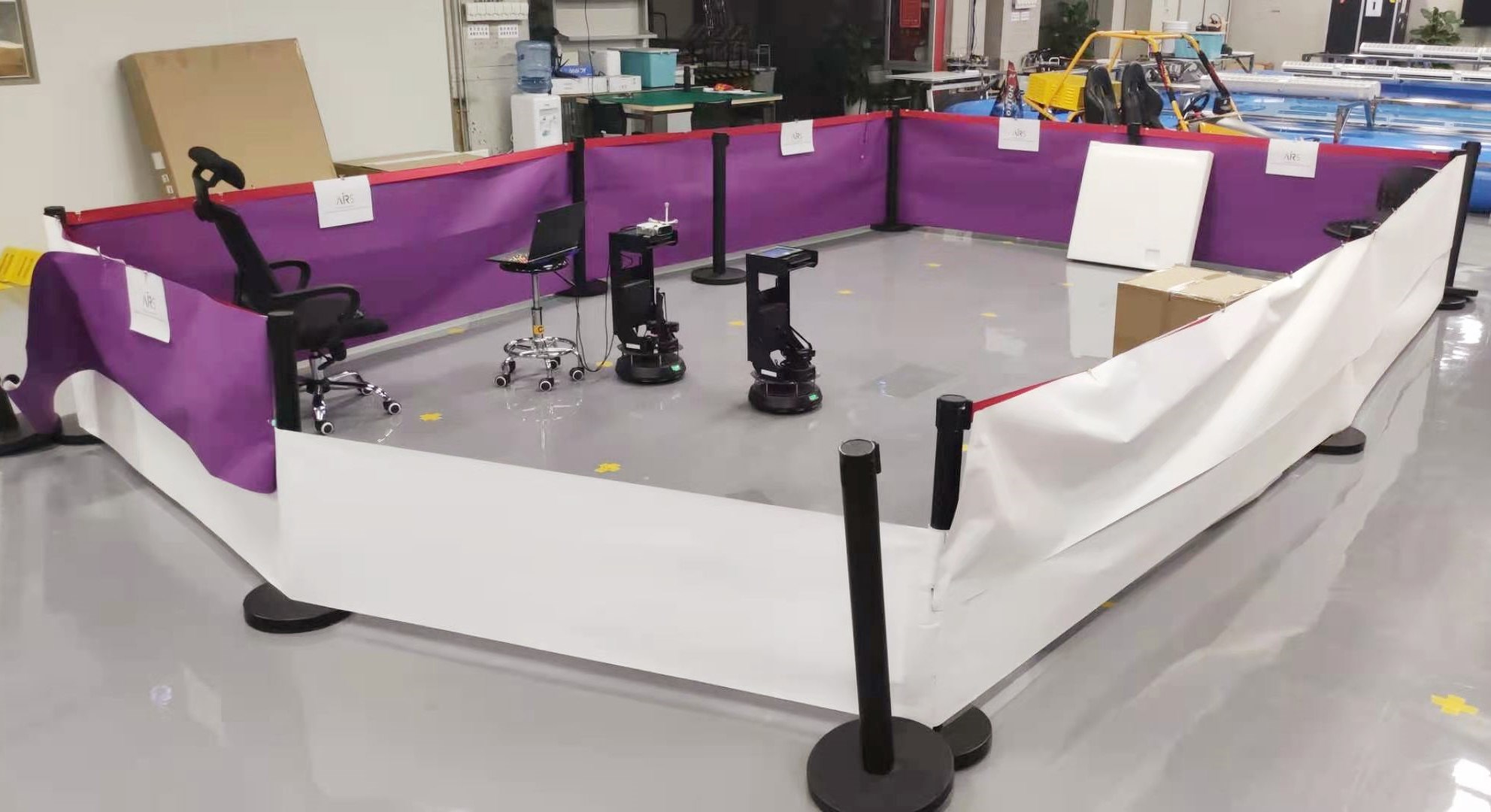} &
    \includegraphics[height=2.8cm]{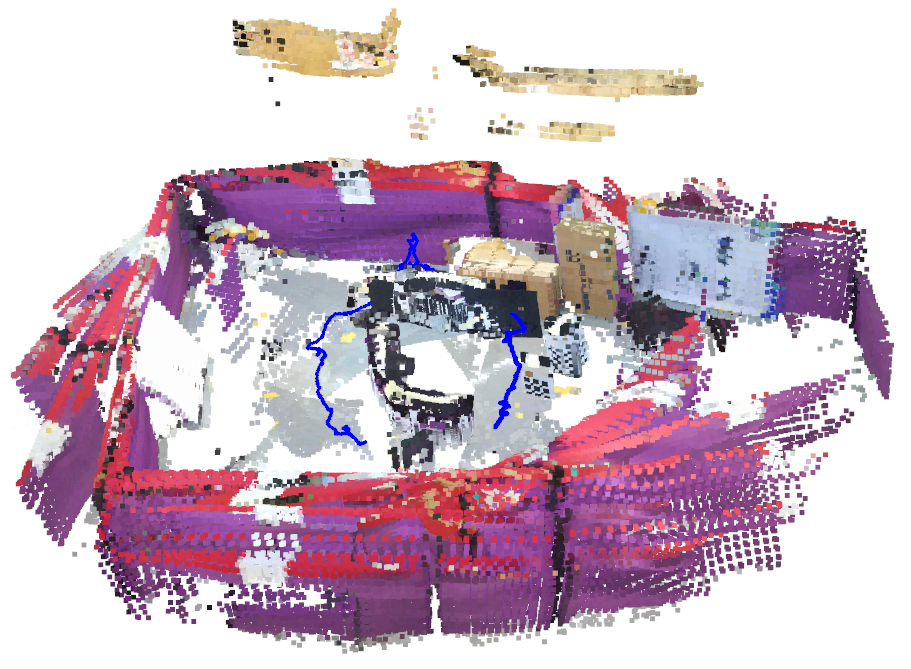} &
    \includegraphics[height=2.4cm]{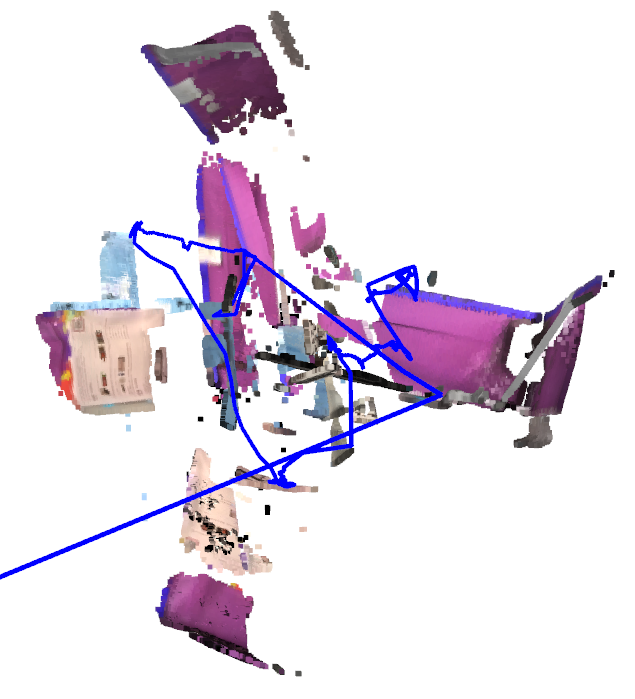} &
    \includegraphics[height=2.8cm]{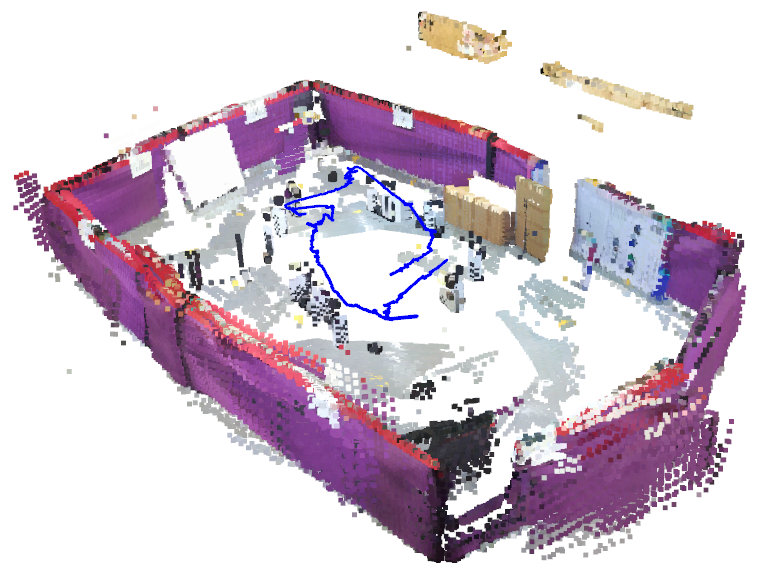} \\
    \includegraphics[height=2.6cm]{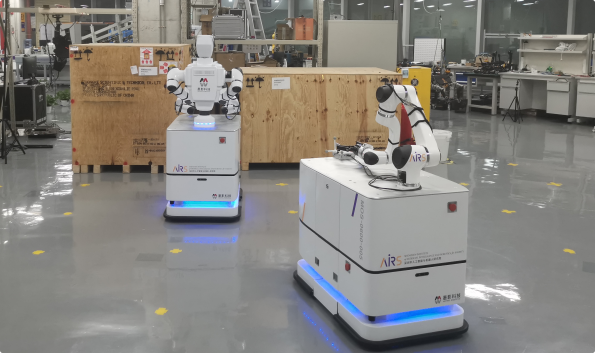} &
    \includegraphics[height=2.8cm]{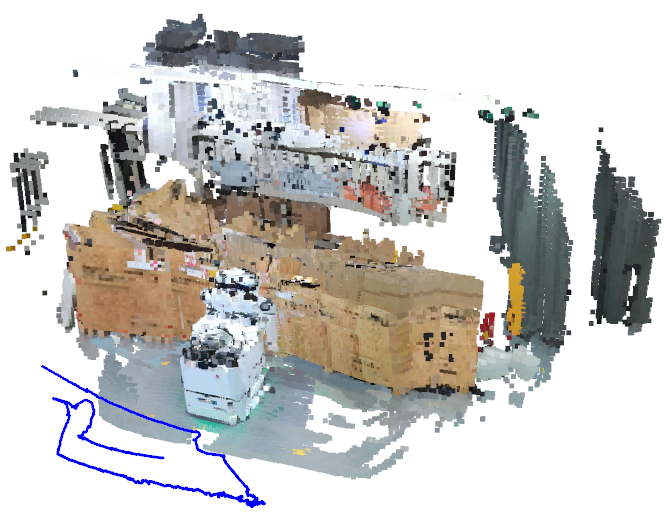} &
    \includegraphics[height=2.4cm]{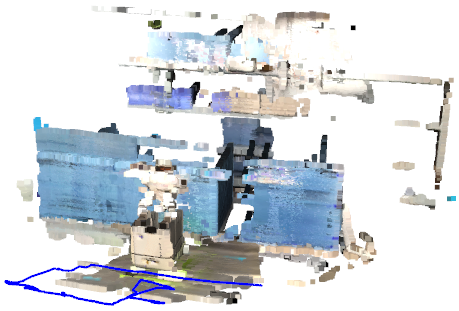} &
    \includegraphics[height=2.8cm]{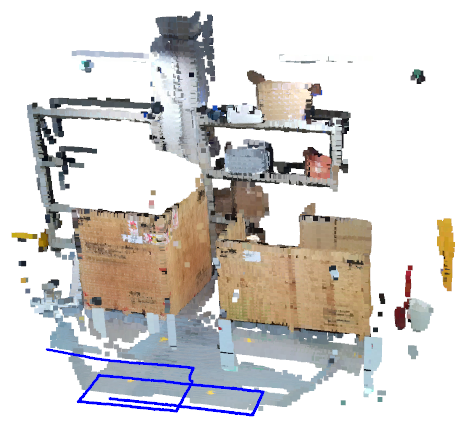} \\
     Sp3 and Mo3 Scenes & ORB2 & SF & Ours \\
  \end{tabular}
  \caption{Dynamic scene mapping results. Top, Sq3 scene. Bottom, Mo3 scene. In both, a robot performs SLAM, another as a dynamic obstacle.  The ORB2 and SF methods fail, while our approach build accurate environment maps. The estimated camera trajectories are displayed in blue.
}
\label{fig:reconstruct}
\vspace{-0.6cm}
\end{figure*}

\subsection{Dynamic SLAM Experiments}
We evaluate the proposed method by comparing the Absolute Trajectory Error (ATE) of the camera trajectory with the original ORB2, ORB2 with sound source object removal (ORB2$+$SSL) and state-of-the-art dense reconstruction dynamic SLAM methods SF in multi-robot dynamic environments.
Sequences starting with ``Sq" and ``Mo" using Spark-T robots and AIRS Dual Arm Mobile Manipulation system robots respectively. The ground truth camera trajectories were obtained from a motion capture system. 

Table \ref{T:ate} lists the ATE Root-Mean-Square-Error (RMSE) of these methods.
The original ORB2 method achieved around 25 $cm$ ATE in the Sp1 and Sp2 sequences, it can be noticed from the camera trajectories that initially it tracked ground truth well in the Sq1 sequence, and after the moving obstacles appeared, the trajectory went wrong. Our method achieved about one-third of the errors of ORB2 in this scene. The ground truth curves are well tracked by our camera trajectories in Fig. \ref{fig:cam-trajectories}. 
SF VO was not robust in these sequences. Its ATE exceeds several meters, which is due to the tendency of its motion segmentation algorithm to judge moving obstacles as static backgrounds and blocks of pixels in the background as dynamic obstacles when large areas are occluded.
This also causes the large and sharp changes in the SF camera trajectories (blue color) and maps in the third image of Fig. \ref{fig:reconstruct}.
ORB2$+$SSL method also obtained small ATE in several sequences, but it's not robust in loop detection. 
Therefore, we chose Openvslam over ORB2 because it has better global loop closure capability, as evaluated in the Sq3 sequence, the proposed approach achieved $8~cm$ less ATE than ORB2$+$SSL, and accurate mapping result (see Fig. \ref{fig:reconstruct}).

For sparse visual feature-based methods like ORB2, the environment map can be synthesized after acquiring the camera trajectory. 
Fig. \ref{fig:reconstruct} shows the final maps produced by the three methods in the Sq3 and Mo3 sequences, where ORB2 distorts the map severely after the presence of obstacles; SF incorrectly segment another robot as a static background, and our method reconstructs the accurate environment map.

Tab. \ref{T:time} compares the online efficiency of several methods. Among them, ORB2 is robust and efficient in static environments and is therefore often used as a base SLAM framework.
DynaSLAM and SF are based on GPU support. DynaSLAM is based on ORB2 and pre-processes dynamic objects using Mask R-CNN with very low frame rates. SF achieves efficient online motion segmentation performance but loses robustness when occlusion is high and is therefore not suitable for multi-robot SLAM. Our approach does not rely on GPU and achieves a 14 \textit{fps} while processing seven microphones with 16,000 Hz sound signal sampling and 125 Hz SSL output.

\section{Discussions}
\textbf{Large-area occlusion} is the biggest problem we encountered in the practical experiments of multi-robot SLAM. When multiple robots approach each other, the images from their cameras are obscured over a large area for the robot behind them. We found that ORB2 vision odometry loses robustness if more than  50\% of the pixels are removed as dynamic objects. This is the reason why the proposed method works better on Spark-T than on the much larger AIRS Dual Arm Mobile robot. 
A similar effect occurs when \textbf{multiple dynamic targets} appear in the visual field at the same time, resulting in large areas of invalid visual features. The introducing an ego-motion prior is a promising approach to cope with such rigid object occlusion problem \cite{rf}.

\section{Conclusions}
\label{sec:conclusion}
In this paper, we have presented a new audio-visual fusion approach that fused SSL into visual SLAM. We apply the SSL results as a dynamic object detector to enable dynamic environment SLAM for mobile robots. 
Experimental results for two different sizes of robots indicate that the proposed method significantly improves the robustness of the visual odometry for the case of severe occlusion in a multi-robot SLAM system. 
In a multi-robot occlusion scenario, the proposed SSL-based SLAM framework achieves real-time performance using a single CPU and outperforms state-of-the-art GPU-based dynamic SLAM solutions.
The future direction of our work is to integrate sound identification into current audio-visual systems for human-robot cooperation.

\section*{ACKNOWLEDGEMENT}
This work is supported by the Shenzhen Institute of Artificial Intelligence and Robotics for Society (2019-ICP002), The Alan Turing Institute and EU H2020 project Enhancing Healthcare with Assistive Robotic Mobile Manipulation (HARMONY, 9911237).
\bibliographystyle{IEEEtran.bst}
\bibliography{IEEEabrv,root}
\end{document}